\documentclass{article}

\usepackage[dblblindworkshop,final]{neurips_2026}
\workshoptitle{On-Device Intelligence: Foundation Models under Real-World Constraints}
\usepackage[utf8]{inputenc}
\usepackage[T1]{fontenc}

\usepackage{hyperref}
\hypersetup{hypertexnames=false,
  pdftitle={EfficientXpert: Efficient Domain Adaptation for Large Language Models via Propagation-Aware Pruning},
  pdfauthor={Songlin Zhao, Michael Pitts, Zhuwei Qin}}
\usepackage{url}
\usepackage{booktabs}
\usepackage{amsfonts}
\usepackage{amsmath}
\usepackage{amssymb}
\usepackage{mathtools}
\usepackage{graphicx}
\usepackage{wrapfig}
\usepackage{algorithm}
\usepackage{algpseudocode}
\usepackage{multirow}
\usepackage{makecell}
\usepackage{microtype}
\usepackage{xcolor}

\title{EfficientXpert: Efficient Domain Adaptation for Large Language Models via Propagation-Aware Pruning}
\author{%
  Songlin Zhao$^1$ \qquad Michael Pitts$^2$ \qquad Zhuwei Qin$^2$ \\
  $^1$University of California, Berkeley \\
  $^2$San Francisco State University \\
  \texttt{tagorezhao@berkeley.edu} \quad \texttt{\{mpitts1,zwqin\}@sfsu.edu}
}

\begin{document}
\maketitle

\begin{abstract}
Deploying domain-specialized large language models on resource-constrained hardware motivates reducing both adaptation cost and the number of retained weights. LoRA lowers adaptation cost but leaves a dense backbone, while separate pruning can discard weights that become important during adaptation. We propose \textbf{EfficientXpert}, a framework that co-adapts low-rank updates and sparse support. Its ForeSight Mask scores weights through a downstream reconstruction surrogate using the evolving LoRA-augmented weights; Partial Brain Surgeon (PBS) realigns the adapter through a closed-form correction. At 40\% sparsity, the strongest variant retains 98.55--108.99\% of dense-LoRA aggregate performance across the evaluated LLaMA settings. On Qwen3-8B, adaptation adds 17.9--27.0\% training time and at most 4.8\% peak GPU memory. Our analysis connects the domain- and rank-dependent effects of PBS to its constrained reconstruction objective, providing guidance for selecting adapter recovery capacity. Together, these results demonstrate efficient production of sparse, domain-specialized experts.
Code is available at \url{https://github.com/TagoreZhao/Efficient_Domain_Adaptation}.
\end{abstract}

\section{Introduction}

Domain-specialized language models support healthcare, law, and finance~\citep{song2025injecting,jin2019pubmedqa,yang2023fingpt,zhou2024lawgpt}, but deploying a separate dense model for each domain is costly on memory- and compute-limited hardware. Low-Rank Adaptation (LoRA) reduces training cost by learning low-rank updates while freezing the pretrained backbone~\citep{hu2022lora}. After merging the adapter, however, the model retains the backbone's dense dimensions. Pruning offers a complementary route to fewer nonzero parameters~\citep{han2015deep,frantar2023sparsegpt,sun2024simpleprune}.

Combining these techniques as independent stages creates a support mismatch. Pruning before adaptation fixes a mask without the learned domain update; reapplying it afterward can discard part of that update. Pruning after adaptation can instead remove coordinates that became domain-critical. Domain-aware methods couple pruning and fine-tuning, but gradient-based importance estimates or auxiliary decision networks introduce additional adaptation machinery~\citep{zhang2024pruning,lu2024allinone}. Producing each sparse expert should therefore be evaluated alongside its retained task performance.

\begin{figure}[t]
  \centering
  \includegraphics[width=\linewidth]{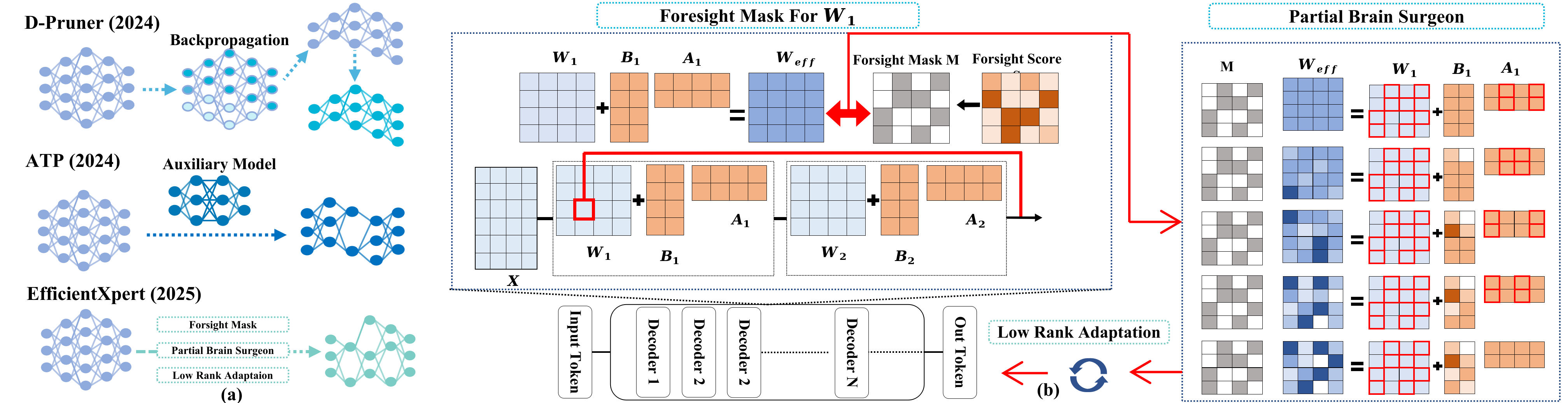}
  \caption{(a) Comparison of EfficientXpert with existing domain-pruning methods. (b) EfficientXpert jointly adapts and prunes a pretrained model: ForeSight uses downstream propagation to update a domain-aware mask, PBS realigns the LoRA adapter, and the final adapter is merged before the mask is applied to produce a sparse expert.}
  \label{fig:overview}
  \vspace{-5mm}
\end{figure}

We propose \textbf{EfficientXpert} (Figure~\ref{fig:overview}), which alternates ordinary LoRA updates with two operations. \textbf{ForeSight Mask} uses a downstream reconstruction surrogate and the evolving effective weights to update sparse support without an additional backward pass. \textbf{Partial Brain Surgeon} solves a weighted ridge problem that suppresses pruned coordinates while controlling changes on retained coordinates. The two components separate support selection from adapter recovery, allowing the correction to be selected according to the target domain and available rank.

Our evaluation spans health and legal domains, four task types, and LLaMA and Qwen3 backbones. At least one variant exceeds post-hoc Wanda's aggregate retention in every evaluated LLaMA setting at 40\% and 50\% sparsity. We quantify the accompanying adaptation overhead and analyze how domain, sparsity, and rank influence the contribution of PBS.

\section{Motivation and related work}
\label{sec:motivation}

SparseGPT and Wanda prune pretrained LLMs using calibration-based reconstruction or activation statistics~\citep{frantar2023sparsegpt,sun2024simpleprune}. LoRA and related parameter-efficient methods address adaptation cost~\citep{hu2022lora,li2021prefix}, while LoRAPrune explores their combination with pruning~\citep{zhang2023loraprune}. Domain adaptation changes both weights and activations, so a pretrained mask need not preserve the resulting specialization. D-Pruner combines general and domain-specific importance signals with gradient-based approximations; ATP learns pruning decisions during adaptation~\citep{zhang2024pruning,lu2024allinone}. EfficientXpert instead uses forward statistics and a small adapter-space correction. Further context appears in Appendix~\ref{app:related-work}.

\begin{wrapfigure}{r}{0.43\linewidth}
  \vspace{-4mm}
  \centering
  \includegraphics[width=\linewidth]{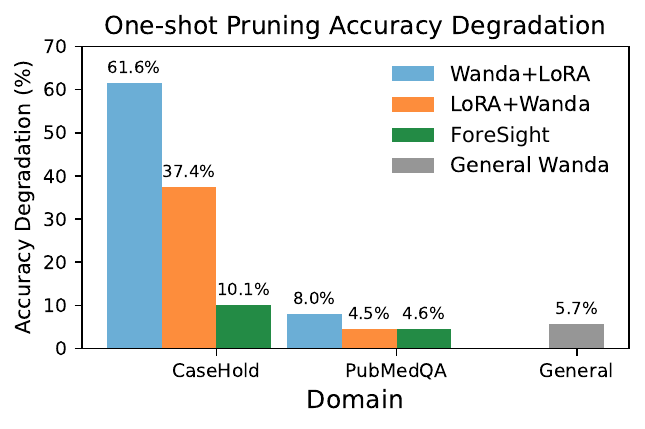}
  \vspace{-4mm}
  \caption{Accuracy degradation at 50\% sparsity on LoRA-adapted LLaMA-2-7B (10-run mean; lower is better) for Wanda+LoRA, LoRA+Wanda, ForeSight, and the published General Wanda mean.}
  \label{fig:acc-degrad}
  \vspace{-5mm}
\end{wrapfigure}
Figure~\ref{fig:acc-degrad} compares sequential pruning and adaptation with ForeSight. On CaseHold, Wanda+LoRA, LoRA+Wanda, and ForeSight degrade accuracy by 61.6\%, 37.4\%, and 10.1\%; the corresponding PubMedQA drops are 8.0\%, 4.5\%, and 4.6\%. ForeSight therefore reduces CaseHold degradation while remaining comparable to post-hoc pruning on PubMedQA. The 5.7\% General Wanda reference is the mean reported in the original Wanda study. The domain contrast motivates tracking adaptation when selecting support. In the main experiments, the post-hoc baseline also receives the fully adapted effective weights. Its comparison with EfficientXpert evaluates the joint procedure, including the evolving support and timing of mask updates.

\WFclear
\section{EfficientXpert}
\label{sec:method}

Consider consecutive effective PyTorch weights $U_1=W_1+B_1A_1\in\mathbb{R}^{h\times d}$ and $U_2=W_2+B_2A_2\in\mathbb{R}^{p\times h}$. Here $B_1\in\mathbb{R}^{h\times r}$, $A_1\in\mathbb{R}^{r\times d}$, and the fixed LoRA scaling is absorbed into the factors. For flattened activations $X\in\mathbb{R}^{\ell\times d}$, the linear output is $XU_1^\top U_2^\top$. We learn a binary mask $M$ for $U_1$ at target sparsity $\rho$.

\subsection{ForeSight Mask: propagation-aware scoring}

Local reconstruction measures $\|X(M\odot U_1-U_1)^\top\|_F$, ignoring the next map. ForeSight uses the downstream surrogate
\begin{equation}
\mathcal{L}_{\mathrm{FS}}(M)=
\left\|X\bigl(M\odot U_1-U_1\bigr)^\top U_2^\top\right\|_F^2.
\label{eq:fs-loss}
\end{equation}
Removing one entry $\theta_{ji}=U_1[j,i]$ creates the outer-product perturbation $-\theta_{ji}(Xe_i)(U_2e_j)^\top$. Its squared norm factorizes exactly:
\begin{equation}
s_{ji}=\theta_{ji}^2
\bigl(X^\top X\bigr)_{ii}
\bigl(U_2^\top U_2\bigr)_{jj}.
\label{eq:fs-score}
\end{equation}
The factors represent weight magnitude, input energy, and downstream amplification. Effective weights incorporate the evolving domain update without an additional gradient objective. The single-coordinate identity requires no stationary-point assumption. Using independent scores to remove many weights neglects interaction terms; attention and gated-MLP extensions additionally use linear-projection surrogates for nonlinear blocks (Appendices~\ref{sec:attention-foresight}--\ref{app:foresight-importance}).

At epoch $t$, we smooth scores as $\bar S_t=\eta S_t+(1-\eta)\bar S_{t-1}$ and retain the largest $(1-\rho)$ fraction per row, with $\eta=0.5$. The downstream factor in Equation~\ref{eq:fs-score} is constant within a row, so a positive factor preserves that row's instantaneous ranking relative to squared Wanda scores on the same weights and activations. Across epochs, changing downstream factors can reweight the score history through EMA. This temporal weighting operates alongside the evolving effective weights, making mask updates responsive to the course of domain adaptation.

\subsection{Partial Brain Surgeon}

With fixed rank, dense LoRA factors can spend capacity on coordinates that the mask removes. Following support selection, PBS updates $B_1$ with $A_1$ fixed. Let $S_i=\{j:M_{ij}=0\}$ and $K_i=\{j:M_{ij}=1\}$, and $D=\operatorname{diag}(\operatorname{diag}(X^\top X))$. For each output row, solve
\begin{equation}
\begin{aligned}
\min_{\Delta b_i}\quad &
\left\|(U_{1,i,S_i}+\Delta b_iA_{1,:,S_i})D_{S_i}^{1/2}\right\|_2^2 \\
&+\left\|\Delta b_iA_{1,:,K_i}D_{K_i}^{1/2}\right\|_2^2
+\lambda\|\Delta b_i\|_2^2.
\end{aligned}
\label{eq:pbs-objective}
\end{equation}
The first term suppresses pruned coordinates; the second penalizes drift on retained coordinates; the third regularizes the correction. Since $S_i$ and $K_i$ partition the columns, their Gram terms sum to $A_1DA_1^\top$, giving
\begin{equation}
\Delta b_i=-U_{1,i,S_i}D_{S_i}A_{1,:,S_i}^{\top}
\left(A_1DA_1^\top+\lambda I_r\right)^{-1}.
\label{eq:pbs}
\end{equation}
One $r\times r$ factorization is shared across all rows; the inverse notation denotes a linear solve. We set $B_1\leftarrow B_1+\Delta B$, with $\lambda=10^{-8}$, then resume LoRA training. This provides an adapter initialization aligned with the current mask. For fixed $A_1$, each correction $\Delta b_iA_1$ lies in its row space, with at most $r$ degrees of freedom. Suppressing the entries in $S_i$ and preserving those in $K_i$ therefore compete for the same adapter capacity. PBS balances these requirements through activation weighting and ridge regularization; subsequent LoRA updates can adapt both factors to the domain objective. This distinction between support selection and constrained recovery explains why their contributions depend on rank and sparsity.

\begin{algorithm}[t]
\caption{EfficientXpert}
\label{alg:EfficientXpert}
\small
\begin{algorithmic}[1]
\Require Backbone $W$, adapters $(B,A)$, training data $\mathbb D$, calibration data $\mathbb D_{\rm cal}$, sparsity $\rho$, epochs $T$, EMA rate $\eta$
\State Initialize $M_0^l\gets\mathbf 1$, $\bar S_0^l\gets0$ for each prunable layer $l$
\For{$t=1,\ldots,T$}
  \State Train LoRA adapters on $\mathbb D$ for one epoch
  \For{each prunable layer $l$}
    \State Compute ForeSight scores $S_t^l$ using $\mathbb D_{\rm cal}$ and effective weights
    \State $\bar S_t^l\gets\eta S_t^l+(1-\eta)\bar S_{t-1}^l$
    \State $M_t^l\gets$ row-wise mask retaining the top $(1-\rho)$ fraction of $\bar S_t^l$
    \State If PBS is enabled, $B^l\gets B^l+\Delta B^l$ using Equation~\ref{eq:pbs}
  \EndFor
\EndFor
\State Merge adapters, then apply $M_T$ to obtain the sparse expert
\end{algorithmic}
\end{algorithm}

\paragraph{Procedure and cost.}
Algorithm~\ref{alg:EfficientXpert} alternates adaptation, mask updates, and optional PBS. Forming an $m\times n$ effective weight from rank-$r$ factors costs $\mathcal O(mnr)$; scoring already formed weights with cached norm factors costs $\mathcal O(mn)$. This weight-formation cost also applies to Wanda when its adapted weights have not been merged. PBS forms and factorizes a shared system and solves its row right-hand sides in $\mathcal O(mnr+mr^2+nr^2+r^3)$ time, with $\mathcal O(mr+r^2)$ workspace when weighted products are accumulated without additional dense copies. Its cost grows with rank; the main experiments use $r=8$.

\section{Experiments}
\label{sec:experiments}

\paragraph{Setup.}
We evaluate LLaMA-2-7B, LLaMA-3.1-8B, and Qwen3 backbones~\citep{touvron2023llama,dubey2024llama,yang2025qwen3}. Each domain mixes 7,000 NLI, 7,000 QA, and 1,000 summarization examples: MedNLI, PubMedQA, and HealthQuestionSummarization for health; ContractNLI, CaseHold, and BillSum for legal~\citep{romanov2018clinicalnli,jin2019pubmedqa,koreeda2021contractnli,zheng2021does,kornilova2019billsum}. Training uses three epochs, rank 8, LoRA scaling parameter 16, learning rate $10^{-4}$, global batch size 16, and sequence length up to 2,048 on four 80-GB NVIDIA A100 GPUs. Calibration uses 128 C4 sequences. Main task results average four runs; the motivation and calibration studies use their separately stated counts. Appendices~\ref{app:dataset-construction}--\ref{app:alpaca-templates} document sampling, hyperparameters, evaluation sets, and prompts.

\paragraph{Baselines and metric.}
Dense+LoRA supplies the unpruned reference. LoRA+Wanda prunes the adapted effective weights $W+BA$, so both methods have access to the domain update. Published D-Pruner and ATP results are reproduced under their reported protocols in Appendix~\ref{app:complete-results}. For complete rows, relative performance is $\frac{100}{7}\sum_{k=1}^{7}q_k^{\rm sparse}/q_k^{\rm dense}$ over NLI and QA accuracy/macro-F1 and summarization ROUGE-1/2/L. Perplexity is reported separately. The aggregate gives equal weight to each of the seven metrics; values above 100\% indicate greater mean normalized performance than dense LoRA. Aggregates are reported only where all seven metrics are available.

\begin{table*}[t]
\centering
\caption{
Main results at 40\% sparsity across two domains and model backbones.
Dense+LoRA denotes the corresponding unpruned baseline. Full comparisons with ATP and D-Pruner and 50\% results are in Appendix~\ref{app:complete-results}.
}
\label{tab:main-results}

\footnotesize
\setlength{\tabcolsep}{2.5pt}
\renewcommand{\arraystretch}{0.94}

\begin{tabular*}{\textwidth}{
    @{\extracolsep{\fill}}lrrrrrrrrr@{}
}
\toprule
& \textbf{PPL}$\downarrow$
& \multicolumn{2}{c}{\textbf{Task 1}}
& \multicolumn{2}{c}{\textbf{Task 2}}
& \multicolumn{3}{c}{\textbf{Summ.}}
& \textbf{Rel.\%}$\uparrow$ \\
\cmidrule(lr){3-4}
\cmidrule(lr){5-6}
\cmidrule(lr){7-9}
\textbf{Method}
& & \textbf{Acc.} & \textbf{F1}
& \textbf{Acc.} & \textbf{F1}
& \textbf{R1} & \textbf{R2} & \textbf{RL}
& \\
\midrule

\multicolumn{10}{c}{
    \textbf{Health -- LLaMA2-7B}
} \\
Dense+LoRA
& 5.53 & 68.05 & 60.22 & 73.16 & 43.98
& 29.52 & 10.63 & 26.77 & 100.00 \\
LoRA+Wanda
& 5.82 & 66.09 & 62.18$^{*}$ & 61.70 & 44.02$^{*}$
& 26.94 & 9.26 & 24.16 & 93.35 \\
ForeSight
& 5.82 & 65.21 & 57.66 & \textbf{71.95} & 47.84$^{*}$
& 29.10 & 10.04 & 26.11 & 98.47 \\
ForeSight+PBS
& \textbf{5.74}
& \textbf{74.51}$^{*}$
& \textbf{74.83}$^{*}$
& \textbf{71.95}
& \textbf{51.08}$^{*}$
& \textbf{30.93}$^{*}$
& \textbf{11.30}$^{*}$
& \textbf{27.74}$^{*}$
& \textbf{108.99} \\

\midrule
\multicolumn{10}{c}{
    \textbf{Health -- LLaMA3.1-8B}
} \\
Dense+LoRA
& 6.49 & 68.28 & 68.17 & 73.60 & 52.12
& 27.98 & 10.51 & 25.04 & 100.00 \\
LoRA+Wanda
& \textbf{7.08}
& \textbf{72.66}$^{*}$
& 54.97 & 73.85$^{*}$ & 52.12
& 26.52 & \textbf{9.95} & 23.22 & 95.65 \\
ForeSight
& \textbf{7.08}
& 69.03$^{*}$ & 63.63
& \textbf{75.00}$^{*}$
& \textbf{52.90}$^{*}$
& 26.32 & 9.42 & 23.45 & 96.45 \\
ForeSight+PBS
& 7.14 & 67.88
& \textbf{68.21}$^{*}$
& 73.92$^{*}$ & 52.24$^{*}$
& \textbf{28.20}$^{*}$
& 9.90
& \textbf{25.45}$^{*}$
& \textbf{99.54} \\

\midrule
\multicolumn{10}{c}{
    \textbf{Legal -- LLaMA2-7B}
} \\
Dense+LoRA
& 3.84 & 75.84 & 72.50 & 70.62 & 68.19
& 32.71 & 13.93 & 18.30 & 100.00 \\
LoRA+Wanda
& 4.09 & 57.62 & 49.90 & 46.12 & 46.35
& 35.33$^{*}$ & 11.83 & 18.67$^{*}$ & 81.86 \\
ForeSight
& 3.96
& \textbf{77.40}$^{*}$
& \textbf{74.99}$^{*}$
& 68.50 & 68.50$^{*}$
& \textbf{37.65}$^{*}$
& \textbf{13.81}
& \textbf{19.94}$^{*}$
& \textbf{103.74} \\
ForeSight+PBS
& \textbf{3.88}
& 74.13 & 70.35
& \textbf{70.50}
& \textbf{70.36}$^{*}$
& 37.47$^{*}$ & 13.61 & 19.91$^{*}$
& 102.69 \\

\midrule
\multicolumn{10}{c}{
    \textbf{Legal -- LLaMA3.1-8B}
} \\
Dense+LoRA
& 4.44 & 76.14 & 74.97 & 82.50 & 68.65
& 33.64 & 15.58 & 19.66 & 100.00 \\
LoRA+Wanda
& 4.83 & 59.46 & 48.30 & 62.50 & 62.28
& 35.54$^{*}$ & 14.05 & 20.16$^{*}$ & 86.77 \\
ForeSight
& 4.85 & 75.60 & 74.29
& \textbf{73.15}
& \textbf{72.94}$^{*}$
& 35.38$^{*}$ & 14.08 & 19.86$^{*}$
& \textbf{98.55} \\
ForeSight+PBS
& \textbf{4.82}
& \textbf{77.05}$^{*}$
& \textbf{74.97}
& 64.00 & 63.48
& \textbf{35.77}$^{*}$
& \textbf{14.18}
& \textbf{20.23}$^{*}$
& 95.93 \\

\bottomrule
\end{tabular*}

\vspace{1mm}
\parbox{\textwidth}{\scriptsize
\textit{Task mapping.}
Health: PPL/Harrison, Task~1/MedNLI, Task~2/PubMedQA,
and Summ./HealthQuestion.
Legal: PPL/MultiLegalPile, Task~1/ContractNLI,
Task~2/CaseHold, and Summ./BillSum.
Bold indicates the best pruned result in each block.
$^{*}$ indicates a strict improvement over the dense baseline on an individual metric at the displayed precision.
PPL is lower-is-better, and all other metrics are higher-is-better.
}
\vspace{-3mm}
\end{table*}

\subsection{Main results and PBS reliability}
\label{sec:ablations}

Table~\ref{tab:main-results} reports 40\% sparsity. The strongest variants achieve 108.99\% and 103.74\% retention on LLaMA-2-7B health and legal, and 99.54\% and 98.55\% on LLaMA-3.1-8B. Corresponding Wanda scores are 93.35\%, 81.86\%, 95.65\%, and 86.77\%. ForeSight+PBS gives the strongest aggregate in health, while ForeSight alone gives the strongest aggregate in legal for both backbones.

\paragraph{When does PBS help?}
At 40\% sparsity, adding PBS changes aggregate retention by $+10.52$ and $+3.09$ percentage points in health for the two backbones, but by $-1.05$ and $-2.62$ points in legal. At 50\%, PBS reduces the aggregate in all four settings: health changes by $-3.45$ and $-12.18$ points, and legal by $-0.55$ and $-2.47$ (Appendix~\ref{app:complete-results}). The trade-off in Equation~\ref{eq:pbs-objective} provides a way to interpret these outcomes: suppression and retained-weight stability use the same low-rank correction. At fixed rank, greater sparsity increases the number of coordinates targeted for suppression. A correction can consequently benefit some tasks while shifting retained weights in ways that reduce others. PBS is most useful when this realignment improves the target-domain objective, motivating selection using the intended task metric.

\paragraph{Rank sensitivity.}
At 50\% sparsity on LLaMA-3.2-1B, ranks 8, 16, 32, and 64 yield 79.57, 78.00, 76.97, and 82.67\% for ForeSight, versus 74.18, 79.28, 77.89, and 82.94\% with PBS (Table~\ref{tab:performance-summary}). Both sequences are non-monotonic, with their highest aggregate at rank 64. Each ratio uses its own rank-matched dense baseline: ForeSight MedNLI accuracy changes from 42.86 at rank 8 to 39.26 at rank 64 even as aggregate retention increases. Increasing rank expands the correction space, while training produces different adapter factors and reference models at each rank. Consequently, the reconstruction capacity and normalized task score can follow different trends. The endpoint gain from rank 8 to 64 is 3.10 points for ForeSight and 8.76 points with PBS, consistent with the latter's greater reliance on adapter recovery capacity. Appendix~\ref{app:PBS_rank} reports all metrics.

\begin{figure}[!t]
  \vspace{-4mm}
  \centering
  \includegraphics[width=\linewidth]{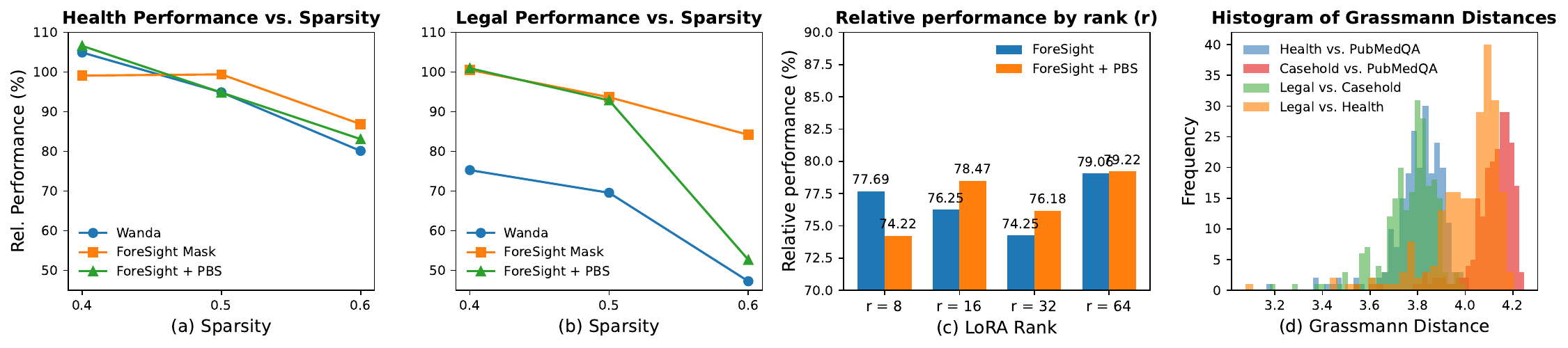}
  \vspace{-2mm}
  \caption{(a--b) LLaMA-2-7B relative performance versus sparsity in the health and legal domains for Wanda, ForeSight, and ForeSight+PBS. (c) Post-pruning relative performance as a function of LoRA rank $r$. (d) Grassmann-distance histograms contrasting task shifts with domain shifts.}
  \label{fig:results}
  \vspace{-1mm}
  \includegraphics[width=\linewidth,trim=8pt 2pt 8pt 2pt,clip]{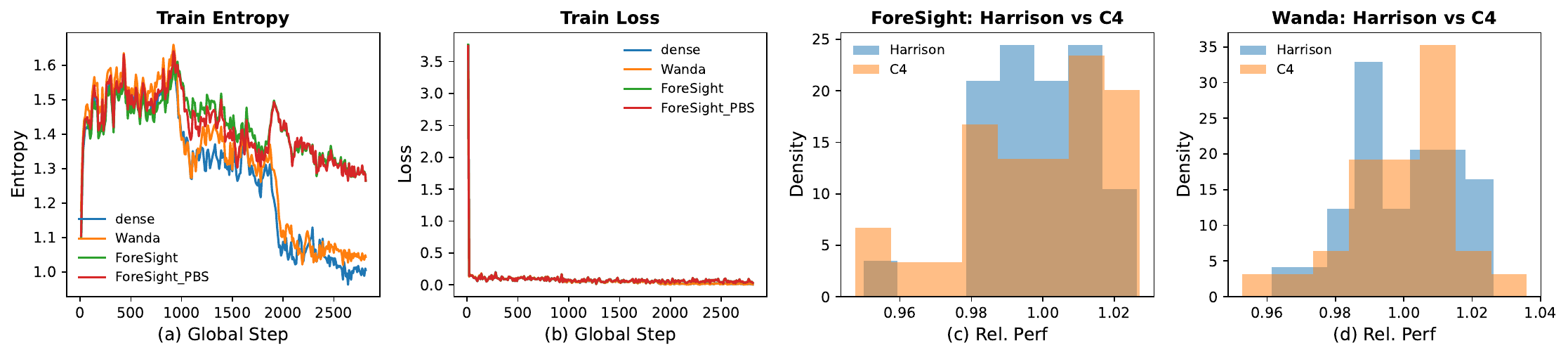}
  \vspace{-2mm}
  \caption{Qwen health diagnostics. (a) Dense LoRA and LoRA+Wanda undergo a late entropy collapse, whereas ForeSight variants maintain higher entropy. (b) Training loss. (c--d) At 30\% sparsity, health-domain Harrison and general-domain C4 calibration produce overlapping relative-performance distributions over 30 seeds for ForeSight and Wanda.}
  \label{fig:qwen-main}
  \vspace{-5mm}
\end{figure}

\paragraph{Practical selection.}
For practical use, we recommend starting with the evaluated rank-8 ForeSight configuration. Compare optional PBS at the same rank and sparsity using a held-out target-domain metric, and enable it when validation improves. Explore higher ranks within the available adaptation budget, assessing absolute task scores alongside normalized retention. This procedure ties the choice of recovery capacity to the application: a summarization expert can prioritize ROUGE, while an NLI or QA expert can prioritize its corresponding accuracy and F1 scores.

\subsection{Adaptation diagnostics}

On Qwen3-8B, ForeSight+PBS follows the dense-LoRA loss trajectory while maintaining higher late-stage output entropy (Figure~\ref{fig:qwen-main}(a--b)). The loss curves show that domain fitting proceeds alongside the mask updates and adapter corrections, while the entropy curves characterize differences in the resulting output distributions.

On Qwen3-0.6B at 30\% sparsity, Harrison and C4 calibration use the same adapted checkpoint, sequence counts, and context lengths. Distributions over 30 seeds overlap for both ForeSight and Wanda (Figure~\ref{fig:qwen-main}(c--d)); domain-matched calibration offers no clear advantage in this setting.

\paragraph{Domain identity shapes adapter geometry.}
Figure~\ref{fig:results}(d) examines whether LoRA's learned directions are organized by domain or task. At matching weight matrices, we compare the eight-dimensional leading left-singular subspaces of the adapter updates. Grassmann distance summarizes their principal angles: larger values indicate more distinct update directions (Appendix~\ref{app:geometry}). Legal--Health and CaseHold--PubMedQA are cross-domain comparisons; Legal--CaseHold and Health--PubMedQA compare a domain mixture with a single task within that domain. Cross-domain pairs concentrate at larger distances, including the two QA-specialized models, while within-domain pairs remain closer despite differing task coverage. This pattern suggests that shared domain content organizes adapter directions beyond task format alone. It motivates selecting sparse support using adapted weights, since pretrained importance can miss directions emphasized during specialization. Together with the calibration study, this supports introducing domain information through evolving adapters while retaining a general calibration corpus.

\subsection{Adaptation efficiency}
\label{sec:efficiency}

Table~\ref{tab:efficiency} measures end-to-end adaptation, including LoRA, masking, and PBS. Legal runtime rises from 106 to 125 minutes ($+17.9\%$), and health from 74 to 94 minutes ($+27.0\%$). Peak allocated GPU memory increases by at most 4.8\%. These measurements quantify the additional training resources needed to obtain a sparse domain expert from the same adaptation pipeline. ForeSight uses forward statistics and effective weights, while PBS reuses a small factorization across the layer's rows. These operations supplement ordinary LoRA training without an auxiliary pruning-decision network or a separate full-gradient saliency objective.

\begin{table}[t]
\vspace{-2mm}
\centering
\caption{End-to-end Qwen-8B adaptation efficiency in the health and legal domains. Runtime includes LoRA training, ForeSight masking, and PBS. Memory is peak allocated GPU memory. $\Delta$Time and $\Delta$Mem report overhead relative to dense LoRA within each domain.}
\label{tab:efficiency}
\begin{minipage}[t]{0.49\textwidth}
\centering
\textbf{(a) Health}\par\smallskip
\resizebox{\linewidth}{!}{%
\begin{tabular}{lrrrrr}
\toprule
Sparsity & Runtime & $\Delta$Time & Peak memory & $\Delta$Mem & PPL$\downarrow$ \\
\midrule
Dense & 74 min & -- & 67,949 MiB & -- & 8.4086 \\
40\%  & 94 min & +27.0\% & 71,239 MiB & +4.8\% & \textbf{8.0113} \\
50\%  & 94 min & +27.0\% & 67,949 MiB & +0.0\% & 8.7747 \\
\bottomrule
\end{tabular}%
}
\end{minipage}\hfill
\begin{minipage}[t]{0.49\textwidth}
\centering
\textbf{(b) Legal}\par\smallskip
\resizebox{\linewidth}{!}{%
\begin{tabular}{lrrrrr}
\toprule
Sparsity & Runtime & $\Delta$Time & Peak memory & $\Delta$Mem & PPL$\downarrow$ \\
\midrule
Dense & 106 min & -- & 45,729 MiB & -- & \textbf{5.1969} \\
40\%  & 125 min & +17.9\% & 46,277 MiB & +1.2\% & 5.4110 \\
50\%  & 125 min & +17.9\% & 45,990 MiB & +0.6\% & 5.8528 \\
\bottomrule
\end{tabular}%
}
\end{minipage}
\vspace{-3mm}
\end{table}

Increasing sparsity from 40\% to 50\% leaves measured training time unchanged while lowering recorded peak memory. Both settings still score the effective weight matrices and train adapters of the same rank, so the target fraction of retained weights and the adaptation workload are distinct controls. Health perplexity changes from 8.0113 to 8.7747, and legal perplexity from 5.4110 to 5.8528. In these runs, 40\% therefore provides stronger language-model retention, whereas 50\% retains fewer weights with the same recorded runtime. After adaptation, merging the adapter and applying the final mask produces a standalone sparse expert.

\section{Conclusion}

EfficientXpert combines ForeSight's evolving sparse support with a closed-form PBS correction for low-rank domain adaptation. Across health and legal settings, the strongest variants retain 98.55--108.99\% of dense-LoRA aggregate performance at 40\% sparsity, with Qwen3-8B adaptation overhead of 17.9--27.0\% in time and at most 4.8\% in peak GPU memory. The task results and adapter geometry motivate selecting support through domain-adapted weights and choosing recovery capacity using target-domain validation. Together, these components produce sparse experts within a LoRA-based adaptation budget.

\clearpage
\begin{ack}
We thank Michael Mahoney, Shuo Yuan, and Zekai Wang for valuable discussions.
\end{ack}

\bibliographystyle{plainnat}
\bibliography{neurips_2026}
\clearpage

\appendix
\section{Related Work}
\label{app:related-work}
\subsection{Parameter-efficient fine-tuning}
As LLMs grow in size and capability, Parameter-Efficient Fine-Tuning (PEFT) has emerged as a vital strategy for adapting pretrained models to domains and downstream tasks~\cite{li2021prefix}. 
Among these, Low-Rank Adaptation (LoRA) has emerged as one of the most widely adopted approaches for adapting LLMs to new domains and tasks.

LoRA injects trainable low-rank matrices into the attention and
feedforward layers of transformer models, enabling efficient adaptation while leaving the original weights unchanged~\cite{hu2022lora}.
LoRA’s simplicity and strong empirical performance have driven its adoption in domains such as health, law, and finance, including FinGPT~\cite{yang2023fingpt}, Med-PaLM~\cite{singhal2023large}, and LawGPT~\cite{zhou2024lawgpt}.
Despite its success, LoRA does not reduce the size or latency of the final model, since the frozen backbone remains fully intact. 
Recent works have begun to explore combinations of LoRA with pruning on general tasks, seeking to preserve LoRA’s training efficiency while improving deployment efficiency~\cite{zhang2023loraprune}. However, the mechanisms by which low-rank updates capture domain- and task-specific information remain largely unexplored.

\subsection{Neural network pruning}
Pruning is a standard compression technique that removes weights while aiming to preserve task performance~\cite{han2015deep,park2020lookahead,cheng2024survey}. 
Recent methods such as SparseGPT~\cite{frantar2023sparsegpt} and Wanda~\cite{sun2024simpleprune} prune pretrained LLMs using importance signals computed from general calibration data, and can retain general benchmark performance even at high sparsity.

Domain adaptation complicates this picture. Fine-tuning reshapes weights and activation statistics, so pruning decisions derived before adaptation can become stale, while post-hoc pruning can remove coordinates that gained importance during adaptation~\cite{zhang2024pruning,lu2024allinone}.

To address these issues, domain adaptive pruning methods such as D-Pruner and ATP update pruning decisions using domain signals during adaptation. 
D-Pruner combines general and domain specific importance scores via gradient based approximations and regularization to preserve domain critical weights~\cite{zhang2024pruning}. 
ATP learns a pruning decision network to update masks during LoRA fine tuning and regularizes adapters to align with the evolving sparse structure~\cite{lu2024allinone}. 
Although effective, both approaches add substantial overhead: gradient based scoring or auxiliary networks increase training time and introduce nontrivial GPU memory cost on top of an already expensive fine tuning pipeline.

\section{An illustration of error propagation}
\label{sec:error-propagation}

\noindent\textbf{Illustrative Example.}
Consider the following matrix multiplication sequence involving two consecutive linear layers:
\[
\underbrace{\begin{bmatrix} 3 & 6 \end{bmatrix}}_{X} 
\cdot \underbrace{\begin{bmatrix} 2 & 4 \\ 2 & 1 \end{bmatrix}^{\!\top}}_{W_1^\top} 
\cdot \underbrace{\begin{bmatrix} 4 & 8 \\ 4 & 1 \end{bmatrix}^{\!\top}}_{W_2^\top}
= \begin{bmatrix} 216 & 132 \end{bmatrix}.
\]

The \textit{local loss} is \(\|X(M\odot W_1-W_1)^\top\|_F^2\), and the \textit{downstream loss} is \(\|X(M\odot W_1-W_1)^\top W_2^\top\|_F^2\), where \(M\) is the binary mask.
Now consider pruning two different weights in \(W_1\). 

Pruning entry \(W_1[1,1] = 2\) (Mask 1) results in a local loss of \(36\) and a downstream loss of \(1152\). In contrast, pruning entry \(W_1[2,2] = 1\) (Mask 2) 
yields the same local loss of \(36\), but a significantly larger downstream loss of \(2340\). This toy example illustrates that pruning decisions which are indistinguishable under local error metrics may lead to dramatically different downstream errors due 
to \emph{error propagation}. As layer depth increases, this effect compounds, motivating the need for pruning strategies that can \emph{foresee} such interactions.

\section{ForeSight for Attention and MLP}
\label{sec:attention-foresight}

ForeSight extends to self-attention by coupling the \emph{query} ($Q$) and \emph{key} ($K$) scores through their propagation norms.  
Throughout, let $X\!\in\!\mathbb{R}^{l\times d}$ denote the flattened input activations of a calibration batch
($l=b\!\times\!s$, where $b$ is the batch size and $s$ the per-sample sequence length).

\paragraph{Weights and adapters.}
\begin{itemize}
    \item Base projections: $W^{Q},W^{K}\!\in\!\mathbb{R}^{d_k\times d}$ and $W^{V}\!\in\!\mathbb{R}^{d_v\times d}$.
    \item LoRA adapters: $B^{Q},B^{K}\!\in\!\mathbb{R}^{d_k\times r}$, $A^{Q},A^{K}\!\in\!\mathbb{R}^{r\times d}$ and
          $B^{V},A^{V}$ for the value path.
\end{itemize}
The effective projections are $\widetilde W^{Q}=W^{Q}+B^{Q}A^{Q}$ and
$\widetilde W^{K}=W^{K}+B^{K}A^{K}$.

\paragraph{Two-layer interaction.}
The attention ``kernel'' observed by the softmax is
\[
    X\,\widetilde W^{Q\!\top}\,\widetilde W^{K}X^{\top}.
\]

\paragraph{Masking the $Q$ projection.}
With a binary mask $M^{Q}\!\in\!\{0,1\}^{d_k\times d}$ and sparsity
constraint $\|\mathbf 1-M^{Q}\|_{0}=k$, ForeSight uses the linear-projection
surrogate
\[
    \min_{M^{Q}} \;
        \bigl\|X\,[\widetilde W^{Q}-M^{Q}\!\odot\!\widetilde W^{Q}]^{\!\top}
              \widetilde W^{K}\bigr\|_{F}^{2}.
\]
Taking the square root of the corresponding single-coordinate score, which
preserves its ranking, gives
\[
    \Delta\mathcal{L}_{ij} \;\propto\;
    \bigl|\widetilde W^{Q}_{ji}\bigr|\,
    \bigl\|\widetilde W^{K}_{j,:}\bigr\|_{2}\,
    \bigl\|X_{:,i}\bigr\|_{2}.
\]

\paragraph{Masking the $K$ projection.}
An analogous problem is solved for $M^{K}\!\in\!\{0,1\}^{d_k\times d}$:
\[
    \min_{M^{K}} \;
        \bigl\|X\,[\widetilde W^{K}-M^{K}\!\odot\!\widetilde W^{K}]^{\!\top}\widetilde W^{Q}\bigr\|_{F}^{2},
    \quad
    \|\mathbf 1-M^{K}\|_{0}=k,
\]
with score
\[
    \Delta\mathcal{L}_{ij} \;\propto\;
    \bigl|\widetilde W^{K}_{ji}\bigr|\,
    \bigl\|\widetilde W^{Q}_{j,:}\bigr\|_{2}\,
    \bigl\|X_{:,i}\bigr\|_{2}.
\]

These propagation-aware scores give one-pass rankings that incorporate the
paired projection's row norms.
\paragraph{Masking the output projection $O$.}
The attention output is projected back to the residual stream by
$\widetilde W^{O}=W^{O}+B^{O}A^{O}\in\mathbb{R}^{d\times d}$.
Its output is consumed by the MLP \emph{input} of both the gate and up projections.
We therefore weight each output coordinate by the average downstream sensitivity
\[
    \nu^{O}_{u}\;:=\;\tfrac{1}{2}\Bigl(\bigl\|\widetilde W^{\text{gate}}_{:,u}\bigr\|_{2}
    +\bigl\|\widetilde W^{\text{up}}_{:,u}\bigr\|_{2}\Bigr),\qquad u\in[d],
\]
and score each entry of $\widetilde W^{O}$ by
\[
    \Delta\mathcal{L}^{O}_{u v}\;\propto\;
    \bigl|\widetilde W^{O}_{u v}\bigr|\,
    \bigl\|X_{:,v}\bigr\|_{2}\,
    \nu^{O}_{u}.
\]
This matches the implementation $\;|\widetilde W^{O}|\odot X_{\text{hat}}\odot \nu^{O}$ (with broadcasting).

\paragraph{Masking MLP gate/up projections.}
For the gated MLP, let
$\widetilde W^{\text{gate}},\widetilde W^{\text{up}}\in\mathbb{R}^{d_{\!ff}\times d}$
and $\widetilde W^{\text{down}}\in\mathbb{R}^{d\times d_{\!ff}}$ denote the effective
(projection + LoRA) weights.
Both gate and up outputs are mapped back by $\widetilde W^{\text{down}}$, so we use the
per-hidden-unit propagation norm
\[
    \nu^{\text{down}}_{h}\;:=\;\bigl\|\widetilde W^{\text{down}}_{:,h}\bigr\|_{2},\qquad h\in[d_{\!ff}].
\]
Using the same linear-projection approximation as the main derivation, ForeSight assigns the importance scores
\[
    \Delta\mathcal{L}^{\text{gate}}_{h i}\;\propto\;
    \bigl|\widetilde W^{\text{gate}}_{h i}\bigr|\,
    \bigl\|X_{:,i}\bigr\|_{2}\,
    \nu^{\text{down}}_{h},
\]
\[
    \Delta\mathcal{L}^{\text{up}}_{h i}\;\propto\;
    \bigl|\widetilde W^{\text{up}}_{h i}\bigr|\,
    \bigl\|X_{:,i}\bigr\|_{2}\,
    \nu^{\text{down}}_{h}.
\]

\section{Derivation of the ForeSight importance score}
\label{app:foresight-importance}

Let $X\in\mathbb R^{\ell\times d}$ be fixed and let $U_1\in\mathbb R^{h\times d}$ and $U_2\in\mathbb R^{p\times h}$ be consecutive effective PyTorch weights. Their linear output is $XU_1^\top U_2^\top$. Pruning $\theta_{ji}=U_1[j,i]$ changes $U_1$ by $\Delta U_1=-\theta_{ji}e_je_i^\top$, so the resulting output perturbation is
\[
X\Delta U_1^\top U_2^\top
=-\theta_{ji}(Xe_i)(U_2e_j)^\top.
\]
Because this is an outer product, its squared Frobenius norm factorizes exactly:
\[
\boxed{
\Delta\mathcal L_{ji}
=\theta_{ji}^2
(X^\top X)_{ii}
(U_2^\top U_2)_{jj}.}
\]
Thus the score combines the squared weight magnitude, input energy, and downstream amplification without requiring a stationary-point or higher-order approximation. For attention and gated MLP blocks, the implementation applies this linear-projection score before the softmax or elementwise nonlinearity; this is the approximation used by ForeSight.

\section{Implementation Notes}
\label{app:algorithm}

Algorithm~\ref{alg:EfficientXpert} in the main text gives the full epoch-wise procedure. ForeSight alone skips the PBS correction while retaining mask updates. In Equations~\ref{eq:pbs-objective}--\ref{eq:pbs}, the LoRA scaling is absorbed into $B$; an implementation storing unscaled factors must account for that scaling when applying the correction. The shared system $A D A^\top+\lambda I_r$ is positive definite for $\lambda>0$ and nonnegative activation energies, allowing a factorized solve without explicitly forming its inverse.

The squared score in Equation~\ref{eq:fs-score} and its square root have identical instantaneous rankings. Smoothing either quantity over epochs is a separate choice: squaring and temporal averaging do not generally commute. Likewise, a positive downstream multiplier constant within one row cannot change its instantaneous row-wise ranking, but time-varying multipliers can change an EMA of score histories. These properties distinguish instantaneous score ordering from the temporal weighting used to update the mask.

\section{Training Dataset Construction}
\label{app:dataset-construction}

The domain training corpora are summarized in
Table~\ref{tab:domain-train-datasets}. For each domain, we stratify the
original training split by label, shuffle it with a fixed seed of 1234, and
subsample it to the desired size. In \textsc{PubMedQA}, we retain only the
\textit{yes} and \textit{no} classes because the released training split
contains no \textit{maybe} examples. For all other datasets, labels with too
few instances are upsampled by sampling with replacement until the target
count is reached. All training data and evaluation scripts follow the Alpaca
template~\cite{dubois2023alpacafarm}; the full prompt formats are provided in
Appendix~\ref{app:alpaca-templates}.

\begin{table*}[t]
\centering
\caption{Training datasets and label distributions. All splits are shuffled
with seed 1234 and use a maximum sequence length of 2048 tokens.}
\label{tab:domain-train-datasets}
\small
\setlength{\tabcolsep}{5pt}
\renewcommand{\arraystretch}{0.95}

\begin{minipage}[t]{0.47\textwidth}
\centering
\textbf{(a) Health domain}\\[2pt]
\begin{tabular}{@{}llr@{}}
\toprule
\textbf{Dataset} & \textbf{Label} & \textbf{Instances} \\
\midrule
MedNLI   & entailment    & 2,333 \\
         & contradiction & 2,333 \\
         & neutral       & 2,334 \\
\midrule
PubMedQA & yes           & 3,500 \\
         & no            & 3,500 \\
         & maybe         & 0 \\
\midrule
HQS      & --            & 1,000 \\
\bottomrule
\end{tabular}
\end{minipage}
\hfill
\begin{minipage}[t]{0.47\textwidth}
\centering
\textbf{(b) Legal domain}\\[2pt]
\begin{tabular}{@{}llr@{}}
\toprule
\textbf{Dataset} & \textbf{Label} & \textbf{Instances} \\
\midrule
ContractNLI & entailment    & 2,450 \\
            & contradiction & 1,400 \\
            & neutral       & 3,150 \\
\midrule
CaseHold    & 0             & 1,400 \\
            & 1             & 1,400 \\
            & 2             & 1,400 \\
            & 3             & 1,400 \\
            & 4             & 1,400 \\
\midrule
BillSum     & --            & 1,000 \\
\bottomrule
\end{tabular}
\end{minipage}
\end{table*}

The task-specific training sets contain 7,000 examples each. For
\textbf{PubMedQA}, we sample 3,500 \textit{yes} and 3,500 \textit{no}
examples; the \textit{maybe} label is excluded because it is unavailable in
the released training split. For \textbf{CaseHold}, we sample 1,400 examples
from each class in \(\{0,1,2,3,4\}\).

\section{Hyperparameter Settings}
\label{sec:training-details}

Table~\ref{tab:training-hyperparams} reports the training hyperparameters for
the dense Health and Legal models and for the task-specific PubMedQA and
CaseHold models. The same settings are used in both domains. All experiments
are conducted on four NVIDIA A100 GPUs with 80\,GB of memory. We adjust the
per-device batch size and gradient accumulation steps to maintain a global
batch size of 16 for both model backbones.

\begin{table}[t]
\centering
\caption{Training hyperparameters shared by the Health and Legal domains.}
\label{tab:training-hyperparams}
\footnotesize
\setlength{\tabcolsep}{3pt}
\renewcommand{\arraystretch}{0.96}
\begin{tabular}{@{}lcc@{}}
\toprule
\textbf{Hyperparameter}
& \textbf{LLaMA2-7B}
& \textbf{LLaMA3.1-8B} \\
\midrule
Learning rate                & \multicolumn{2}{c}{$1\times10^{-4}$} \\
Per-device batch size        & 2 & 1 \\
Gradient accumulation steps  & 2 & 4 \\
Training epochs              & \multicolumn{2}{c}{3} \\
Weight decay                 & \multicolumn{2}{c}{0.01} \\
FP16                         & \multicolumn{2}{c}{True} \\
Save interval                & \multicolumn{2}{c}{1,000 steps} \\
Maximum sequence length      & \multicolumn{2}{c}{2,048} \\
LoRA rank                    & \multicolumn{2}{c}{8} \\
LoRA scaling parameter       & \multicolumn{2}{c}{16} \\
LoRA dropout                 & \multicolumn{2}{c}{0} \\
LoRA bias                    & \multicolumn{2}{c}{none} \\
Optimizer                    & \multicolumn{2}{c}{\texttt{torch\_Adam}} \\
\bottomrule
\end{tabular}

\vspace{1mm}
\parbox{\columnwidth}{\scriptsize
The LoRA target modules are
\texttt{q\_proj}, \texttt{k\_proj}, \texttt{v\_proj},
\texttt{o\_proj}, \texttt{gate\_proj}, \texttt{up\_proj}, and
\texttt{down\_proj}.
}
\end{table}

The LoRA training hyperparameters and datasets used for EfficientXpert are
identical to those used for the dense models to ensure a fair comparison. For
ForeSight Mask, the mask learning rate is set to 0.5. For Partial Brain
Surgeon, the regularization coefficient is set to \(1\times10^{-8}\).

\section{Evaluation Details}
\label{sec:eval-details}

The evaluation datasets, test-set sizes, tasks, and metrics are summarized in
Table~\ref{tab:evaluation-datasets}. For all evaluations, we use the Alpaca
template~\cite{dubois2023alpacafarm}. The complete prompt templates are
provided in Appendix~\ref{app:alpaca-templates}.

\begin{table*}[t]
\centering
\caption{Evaluation datasets, test-set sizes, and metrics for the Health and
Legal domains. Acc. denotes accuracy and F1 denotes macro-F1. Dataset sources
include MultiLegalPile~\citep{niklaus2023multilegalpile} and
ContractNLI~\citep{koreeda2021contractnli}.}
\label{tab:evaluation-datasets}
\label{tab:health-dataset-summary}
\label{tab:legal-dataset-summary}
\footnotesize
\setlength{\tabcolsep}{4pt}
\renewcommand{\arraystretch}{1.0}

\begin{minipage}[t]{0.48\textwidth}
\centering
\textbf{(a) Health domain}\\[2pt]
\begin{tabular}{@{}llrl@{}}
\toprule
\textbf{Dataset} & \textbf{Task} & \textbf{Test} & \textbf{Metric} \\
\midrule
InternalMed-Harrison & Generation & 300   & PPL \\
MedNLI               & NLI        & 1,422 & Acc., F1 \\
PubMedQA              & QA         & 500   & Acc., F1 \\
HQS                   & Summ.      & 100   & ROUGE \\
\bottomrule
\end{tabular}
\end{minipage}
\hfill
\begin{minipage}[t]{0.48\textwidth}
\centering
\textbf{(b) Legal domain}\\[2pt]
\begin{tabular}{@{}llrl@{}}
\toprule
\textbf{Dataset} & \textbf{Task} & \textbf{Test} & \textbf{Metric} \\
\midrule
MultiLegalPile & Generation & 300   & PPL \\
ContractNLI    & NLI        & 1,991 & Acc., F1 \\
CaseHOLD       & QA         & 200   & Acc., F1 \\
BillSum        & Summ.      & 200   & ROUGE \\
\bottomrule
\end{tabular}
\end{minipage}
\end{table*}

\section{Alpaca Templates}
\label{app:alpaca-templates}

\subsection*{PubMedQA}
\begin{quote}\ttfamily\small\raggedright
Below is an instruction that describes a task related to healthcare, paired with detailed input from a scientific article.

Instruction: You are an expert clinician. Determine whether the following statement is correct. Answer yes, no, or maybe.

Input:
Contexts:
\{contexts\}

Section Labels: \{labels\}

MeSH Terms: \{meshes\}

Question: \{question\}

Response: The answer is
\end{quote}

\subsection*{MedNLI}
\begin{quote}\ttfamily\small\raggedright
Below is an instruction for analyzing clinical information.

Instruction: You are a licensed physician. Determine the logical relationship between two clinical statements based only on clinical knowledge and the given premise. Do not assume unstated facts.

Answer with exactly one label:
- entailment: the hypothesis must be true given the premise
- contradiction: the hypothesis must be false given the premise
- neutral: the hypothesis may be true or false given the premise

Input:
Premise: \{sentence1\}
Hypothesis: \{sentence2\}

Response: The answer is
\end{quote}

\subsection*{HQS}
\begin{quote}\ttfamily\small\raggedright
Below is an instruction that describes a task related to healthcare, paired with further context.

Instruction: Summarize the following consumer health question into a shorter version that preserves all critical information needed to answer it correctly. Keep key entities (e.g., conditions, treatments, tests) and the main intent. Remove peripheral details. Write a fluent sentence.

Input: \{input\_question\}

Response: The shortened question is
\end{quote}

\subsection*{CaseHold}
\begin{quote}\ttfamily\small\raggedright
Below is an instruction that describes a task related to legal reasoning.

Instruction: You are given a citing passage from a legal decision and five candidate holding statements. Select the holding statement that best matches the citing passage. Focus on the specific legal rule, factual framing, and implication. Avoid choices that are merely topically similar.

Answer with a single integer in \{0,1,2,3,4\}.

Input:
Citing passage: \{citing\_prompt\}

0: \{holding\_0\}
1: \{holding\_1\}
2: \{holding\_2\}
3: \{holding\_3\}
4: \{holding\_4\}

Response: The answer is
\end{quote}

\subsection*{ContractNLI}
\begin{quote}\ttfamily\small\raggedright
Below is an instruction that describes a task related to contract analysis.

Instruction: Determine the relationship between the contract premise and the hypothesis. Answer with exactly one label: entailment, contradiction, or neutral. Consider all three options and base your decision only on the provided text.

Input:
Premise: \{sentence1\}
Hypothesis: \{sentence2\}

Response: The answer is
\end{quote}

\subsection*{BillSum}
\begin{quote}\ttfamily\small\raggedright
Below is an instruction for summarizing a legislative bill.

Instruction: Summarize the following bill by explaining its major actions, purposes, and effects. Focus on what the bill aims to do and what changes it makes in practice. Paraphrase clearly for a general policy audience. Avoid quoting long spans or referencing subsection numbers unless needed.

Input:
Bill Title: \{title\}

Bill Text:
\{input\_text\}

Response:
\end{quote}

\section{Complete Domain Results}
\label{app:complete-results}

\begin{table*}[t]
\centering
\caption{Complete results in the \textbf{Health} domain. Dense+LoRA is the
shared unpruned baseline for both sparsity levels.}
\label{tab:appendix-health}
\footnotesize
\setlength{\tabcolsep}{2.5pt}
\renewcommand{\arraystretch}{0.94}

\begin{tabular*}{\textwidth}{@{\extracolsep{\fill}}lrrrrrrrrr@{}}
\toprule
& \textbf{Harrison}
& \multicolumn{2}{c}{\textbf{MedNLI}}
& \multicolumn{2}{c}{\textbf{PubMedQA}}
& \multicolumn{3}{c}{\textbf{HealthQuestion}}
& \textbf{Rel.\%}$\uparrow$ \\
\cmidrule(lr){3-4}\cmidrule(lr){5-6}\cmidrule(lr){7-9}
\textbf{Method}
& \textbf{PPL}$\downarrow$
& \textbf{Acc.} & \textbf{F1}
& \textbf{Acc.} & \textbf{F1}
& \textbf{R1} & \textbf{R2} & \textbf{RL}
& \\
\midrule

\multicolumn{10}{c}{\textbf{LLaMA2-7B}} \\
Dense+LoRA
& 5.53 & 68.05 & 60.22 & 73.16 & 43.98
& 29.52 & 10.63 & 26.77 & 100.00 \\
\addlinespace[1pt]
\multicolumn{10}{l}{\emph{Sparsity 0.5}} \\
D-Pruner
& 6.74 & 61.88 & -- & -- & 32.58
& 36.49 & 13.71 & 31.85 & -- \\
ATP
& 9.53 & 70.51 & -- & -- & 42.06
& 29.66 & 10.36 & 27.38 & -- \\
LoRA+Wanda
& 6.22 & 64.25 & 41.46 & 64.60 & 47.31$^{*}$
& 25.86 & \textbf{15.69}$^{*}$ & 23.19 & 97.28 \\
ForeSight
& \textbf{6.21} & \textbf{72.86}$^{*}$ & 54.48
& \textbf{69.65} & \textbf{49.30}$^{*}$
& 27.81 & 9.27 & \textbf{25.52} & \textbf{97.37} \\
ForeSight+PBS
& \textbf{6.21} & 59.70 & \textbf{58.21}
& 66.40 & 47.22$^{*}$
& \textbf{27.92} & 9.38 & 24.65 & 93.92 \\
\addlinespace[1pt]
\multicolumn{10}{l}{\emph{Sparsity 0.4}} \\
ATP
& 8.47 & 71.52 & -- & -- & 44.60
& 31.33 & 11.15 & 28.21 & -- \\
LoRA+Wanda
& 5.82 & 66.09 & 62.18$^{*}$ & 61.70 & 44.02$^{*}$
& 26.94 & 9.26 & 24.16 & 93.35 \\
ForeSight
& 5.82 & 65.21 & 57.66 & \textbf{71.95} & 47.84$^{*}$
& 29.10 & 10.04 & 26.11 & 98.47 \\
ForeSight+PBS
& \textbf{5.74} & \textbf{74.51}$^{*}$ & \textbf{74.83}$^{*}$
& \textbf{71.95} & \textbf{51.08}$^{*}$
& \textbf{30.93}$^{*}$ & \textbf{11.30}$^{*}$
& \textbf{27.74}$^{*}$ & \textbf{108.99} \\

\midrule
\multicolumn{10}{c}{\textbf{LLaMA3.1-8B}} \\
Dense+LoRA
& 6.49 & 68.28 & 68.17 & 73.60 & 52.12
& 27.98 & 10.51 & 25.04 & 100.00 \\
\addlinespace[1pt]
\multicolumn{10}{l}{\emph{Sparsity 0.5}} \\
ATP
& 13.94 & 68.57 & -- & -- & 36.72
& 27.91 & 9.70 & 24.71 & -- \\
LoRA+Wanda
& 7.93 & 74.77$^{*}$ & 56.36 & 71.70 & 50.94
& 26.40 & 9.61 & 23.34 & 95.19 \\
ForeSight
& \textbf{7.92} & \textbf{76.09}$^{*}$ & \textbf{76.46}$^{*}$
& \textbf{71.95} & \textbf{51.05}
& \textbf{27.84} & \textbf{10.13} & \textbf{23.99}
& \textbf{101.57} \\
ForeSight+PBS
& 8.16 & 57.42 & 56.34 & 70.55 & 49.53
& 25.80 & 8.85 & 22.95 & 89.39 \\
\addlinespace[1pt]
\multicolumn{10}{l}{\emph{Sparsity 0.4}} \\
ATP
& 12.48 & 75.32 & -- & -- & 43.09
& 29.86 & 11.16 & 27.01 & -- \\
LoRA+Wanda
& \textbf{7.08} & \textbf{72.66}$^{*}$ & 54.97
& 73.85$^{*}$ & 52.12
& 26.52 & \textbf{9.95} & 23.22 & 95.65 \\
ForeSight
& \textbf{7.08} & 69.03$^{*}$ & 63.63
& \textbf{75.00}$^{*}$ & \textbf{52.90}$^{*}$
& 26.32 & 9.42 & 23.45 & 96.45 \\
ForeSight+PBS
& 7.14 & 67.88 & \textbf{68.21}$^{*}$
& 73.92$^{*}$ & 52.24$^{*}$
& \textbf{28.20}$^{*}$ & 9.90
& \textbf{25.45}$^{*}$ & \textbf{99.54} \\

\bottomrule
\end{tabular*}

\vspace{1mm}
\parbox{\textwidth}{\scriptsize
Bold marks the best result among LoRA+Wanda, ForeSight, and ForeSight+PBS
within each model and sparsity setting; within those rows, $^{*}$ marks a strict improvement
over the corresponding dense baseline on an individual metric at the displayed
precision. PPL is lower-is-better; all other metrics are higher-is-better. A
dash denotes an unreported result or an aggregate that cannot be computed.
}
\end{table*}

\clearpage

\begin{table*}[t]
\centering
\caption{Complete results in the \textbf{Legal} domain. Dense+LoRA is the
shared unpruned baseline for both sparsity levels.}
\label{tab:appendix-legal}
\footnotesize
\setlength{\tabcolsep}{2.5pt}
\renewcommand{\arraystretch}{0.94}

\begin{tabular*}{\textwidth}{@{\extracolsep{\fill}}lrrrrrrrrr@{}}
\toprule
& \textbf{MultiLegalPile}
& \multicolumn{2}{c}{\textbf{ContractNLI}}
& \multicolumn{2}{c}{\textbf{CaseHold}}
& \multicolumn{3}{c}{\textbf{BillSum}}
& \textbf{Rel.\%}$\uparrow$ \\
\cmidrule(lr){3-4}\cmidrule(lr){5-6}\cmidrule(lr){7-9}
\textbf{Method}
& \textbf{PPL}$\downarrow$
& \textbf{Acc.} & \textbf{F1}
& \textbf{Acc.} & \textbf{F1}
& \textbf{R1} & \textbf{R2} & \textbf{RL}
& \\
\midrule

\multicolumn{10}{c}{\textbf{LLaMA2-7B}} \\
Dense+LoRA
& 3.84 & 75.84 & 72.50 & 70.62 & 68.19
& 32.71 & 13.93 & 18.30 & 100.00 \\
\addlinespace[1pt]
\multicolumn{10}{l}{\emph{Sparsity 0.5}} \\
D-Pruner
& 2.73 & -- & -- & -- & 27.58
& 31.00 & 19.03 & 25.96 & -- \\
ATP
& 3.67 & -- & -- & -- & --
& 43.18 & 23.12 & 30.06 & -- \\
LoRA+Wanda
& 4.35 & 53.60 & 46.73 & 41.33 & 41.15
& 33.31$^{*}$ & 11.50 & 17.79 & 76.51 \\
ForeSight
& 4.10 & \textbf{71.37} & \textbf{68.38}
& \textbf{63.50} & 63.46
& \textbf{34.11}$^{*}$ & 11.94
& \textbf{18.43}$^{*}$ & \textbf{94.59} \\
ForeSight+PBS
& \textbf{4.02} & 69.61 & 67.12
& \textbf{63.50} & \textbf{63.49}
& 33.94$^{*}$ & \textbf{12.04}
& \textbf{18.43}$^{*}$ & 94.04 \\
\addlinespace[1pt]
\multicolumn{10}{l}{\emph{Sparsity 0.4}} \\
ATP
& 2.82 & -- & -- & -- & --
& 43.80 & 23.12 & 30.06 & -- \\
LoRA+Wanda
& 4.09 & 57.62 & 49.90 & 46.12 & 46.35
& 35.33$^{*}$ & 11.83 & 18.67$^{*}$ & 81.86 \\
ForeSight
& 3.96 & \textbf{77.40}$^{*}$ & \textbf{74.99}$^{*}$
& 68.50 & 68.50$^{*}$
& \textbf{37.65}$^{*}$ & \textbf{13.81}
& \textbf{19.94}$^{*}$ & \textbf{103.74} \\
ForeSight+PBS
& \textbf{3.88} & 74.13 & 70.35
& \textbf{70.50} & \textbf{70.36}$^{*}$
& 37.47$^{*}$ & 13.61 & 19.91$^{*}$
& 102.69 \\

\midrule
\multicolumn{10}{c}{\textbf{LLaMA3.1-8B}} \\
Dense+LoRA
& 4.44 & 76.14 & 74.97 & 82.50 & 68.65
& 33.64 & 15.58 & 19.66 & 100.00 \\
\addlinespace[1pt]
\multicolumn{10}{l}{\emph{Sparsity 0.5}} \\
ATP
& 4.28 & -- & -- & -- & --
& 43.10 & 22.60 & 28.65 & -- \\
LoRA+Wanda
& 5.35 & 50.18 & 39.60 & 41.50 & 41.58
& 35.05$^{*}$ & 12.83 & 19.11 & 73.33 \\
ForeSight
& \textbf{5.32} & 72.37 & 71.27 & 67.50 & 67.06
& \textbf{37.81}$^{*}$ & \textbf{14.62}
& \textbf{20.56}$^{*}$ & \textbf{97.20} \\
ForeSight+PBS
& 5.38 & \textbf{74.38} & \textbf{71.94}
& \textbf{68.00} & \textbf{67.71}
& 35.03$^{*}$ & 13.24 & 19.52 & 94.73 \\
\addlinespace[1pt]
\multicolumn{10}{l}{\emph{Sparsity 0.4}} \\
ATP
& 4.13 & -- & -- & -- & --
& 45.28 & 25.45 & 30.15 & -- \\
LoRA+Wanda
& 4.83 & 59.46 & 48.30 & 62.50 & 62.28
& 35.54$^{*}$ & 14.05 & 20.16$^{*}$ & 86.77 \\
ForeSight
& 4.85 & 75.60 & 74.29
& \textbf{73.15} & \textbf{72.94}$^{*}$
& 35.38$^{*}$ & 14.08 & 19.86$^{*}$
& \textbf{98.55} \\
ForeSight+PBS
& \textbf{4.82} & \textbf{77.05}$^{*}$ & \textbf{74.97}
& 64.00 & 63.48
& \textbf{35.77}$^{*}$ & \textbf{14.18}
& \textbf{20.23}$^{*}$ & 95.93 \\

\bottomrule
\end{tabular*}

\vspace{1mm}
\parbox{\textwidth}{\scriptsize
Bold marks the best result among LoRA+Wanda, ForeSight, and ForeSight+PBS
within each model and sparsity setting; within those rows, $^{*}$ marks a strict improvement
over the corresponding dense baseline on an individual metric at the displayed
precision. PPL is lower-is-better; all other metrics are higher-is-better. A
dash denotes an unreported result or an aggregate that cannot be computed.
}
\end{table*}

\clearpage
\section{Effect of Rank \texorpdfstring{\(r\)}{r} on Partial Brain Surgeon}
\label{app:PBS_rank}

PBS has only $r$ correction coefficients per row. Increasing rank expands its capacity, but need not improve task performance; the ridge problem remains well-defined at every rank. Each rank's relative performance uses its corresponding dense baseline.

We evaluated the impact of varying \(r\) at 50\% sparsity on LLaMA-3.2-1B. Figure~\ref{fig:results}(c) shows that both methods achieve their highest aggregate at rank 64, but the intermediate values are non-monotonic. Table~\ref{tab:performance-summary} reports every individual metric.

\begin{table}[!ht]
\vspace{-3mm}
\centering
\caption{Rank sensitivity at 50\% sparsity on LLaMA-3.2-1B. Task scores
are reported in their displayed units, and the aggregate is a percentage.
Bold marks the best pruned task score in each metric column; for Rel.\%, it
marks the best rank within each method. Lower PPL is better, while all other
metrics are higher-is-better.}
\label{tab:performance-summary}
\footnotesize
\setlength{\tabcolsep}{3pt}
\renewcommand{\arraystretch}{0.96}

\begin{tabular*}{\textwidth}{@{\extracolsep{\fill}}lrrrrrrrrrr@{}}
\toprule
& & \textbf{Med}
& \multicolumn{2}{c}{\textbf{MedNLI}}
& \multicolumn{2}{c}{\textbf{PubMedQA}}
& \multicolumn{3}{c}{\textbf{HQS}}
& \textbf{Rel.\%} \\
\cmidrule(lr){4-5}\cmidrule(lr){6-7}\cmidrule(lr){8-10}
\textbf{Method} & \textbf{Rank}
& \textbf{PPL}$\downarrow$
& \textbf{Acc.} & \textbf{F1}
& \textbf{Acc.} & \textbf{F1}
& \textbf{R1} & \textbf{R2} & \textbf{RL}
& $\uparrow$ \\
\midrule

\multirow{4}{*}{Dense+LoRA}
& 8  & 10.288 & 56.05 & 54.64 & 68.00 & 47.14 & 24.76 & 8.96 & 22.04 & 100.00 \\
& 16 & 10.145 & 59.56 & 57.59 & 66.00 & 46.74 & 26.09 & 8.52 & 23.43 & 100.00 \\
& 32 & 10.029 & 60.69 & 59.86 & 66.60 & 46.56 & 25.07 & 8.09 & 22.84 & 100.00 \\
& 64 &  9.902 & 58.72 & 57.33 & 65.00 & 45.75 & 24.39 & 7.55 & 21.22 & 100.00 \\

\midrule
\multirow{4}{*}{ForeSight}
& 8  & 14.716 & 42.86 & 39.61 & 52.00 & 37.10 & 21.71 & 6.88 & 19.48 & 79.57 \\
& 16 & 14.706 & 42.86 & 39.61 & 52.00 & 37.10 & 21.71 & 6.88 & 19.48 & 78.00 \\
& 32 & \textbf{14.685} & 41.97 & 36.20 & 52.05 & 37.16 & 21.14 & 6.72 & 19.14 & 76.97 \\
& 64 & 14.834 & 39.26 & 39.21 & 53.00 & 39.72 & 22.01 & \textbf{6.93} & 19.74 & \textbf{82.67} \\

\midrule
\multirow{4}{*}{ForeSight+PBS}
& 8  & 15.854 & 36.43 & 34.12 & \textbf{57.20} & \textbf{40.27} & 19.60 & 5.68 & 17.57 & 74.18 \\
& 16 & 15.020 & \textbf{43.65} & \textbf{42.52} & 53.95 & 38.45 & 22.06 & 6.37 & 19.80 & 79.28 \\
& 32 & 14.908 & 39.68 & 39.79 & 55.05 & 39.17 & 21.17 & 6.41 & 18.94 & 77.89 \\
& 64 & 15.106 & 39.54 & 38.24 & 54.90 & 39.10 & \textbf{22.27} & 6.88 & \textbf{19.98} & \textbf{82.94} \\

\bottomrule
\end{tabular*}
\vspace{-2mm}
\end{table}

\section{Adapter Geometry}
\label{app:geometry}

For two adapters attached to the same weight matrix, write their effective updates as $\Delta W_a=B_aA_a$ and $\Delta W_b=B_bA_b$. Let $Q_a$ and $Q_b$ contain the leading $r=8$ left singular vectors of these updates, with orthonormal columns. If $\sigma_k$ are the singular values of $Q_a^\top Q_b$, the principal angles are $\theta_k=\arccos(\sigma_k)$ and the Grassmann distance is
\begin{equation}
d_{\mathrm G}(Q_a,Q_b)=\left(\sum_{k=1}^{8}\theta_k^2\right)^{1/2}.
\label{eq:grassmann-distance}
\end{equation}
This distance compares subspace orientation independently of the choice of orthonormal basis. Identical subspaces have distance zero; larger values indicate greater angular separation between the learned update directions.

Figure~\ref{fig:results}(d) summarizes these distances at matching weights. Legal--Health compares two domain mixtures, and CaseHold--PubMedQA compares two QA-specialized adapters across domains. Legal--CaseHold and Health--PubMedQA compare each domain mixture with its corresponding QA task. The larger distances for cross-domain pairs, together with the closer within-domain pairs, suggest that domain identity shapes the directions emphasized by adaptation across different task compositions. This provides a geometric motivation for updating sparse support using the evolving domain-adapted weights.

\end{document}